\documentclass[review]{elsarticle}
\usepackage[utf8]{inputenc}

\usepackage{graphicx}
\usepackage{lineno,hyperref}
\modulolinenumbers[5]

\usepackage
{todonotes}
\usepackage{mathtools}
\usepackage{amsfonts}
\usepackage{amssymb}
\usepackage{amsmath}
\usepackage{wrapfig}
\usepackage{times}
\usepackage{url}
\usepackage[margin=17pt]{subcaption}
\usepackage[final]{changes}
\usepackage{color}
\usepackage{xspace}
\usepackage[shortlabels]{enumitem}
\usepackage{makecell}
\usepackage{colortbl}
\definecolor{LightCyan}{rgb}{0.9,1,1}
\definecolor{LightGray}{rgb}{0.95,0.95,0.95}
\usepackage{rotating}
\newcommand{\ME}{E}

\newcommand{\MW}{\mathcal{W}}
\newcommand{\MT}{\mathcal{T}}
\newcommand{\ML}{\mathcal{L}}

\newcommand{\RR}{\mathbb{R}}
\newcommand{\NN}{\mathbb{N}}

\newtheorem{definition}{Definition}[section]

\newtheorem{example}{Example}[section]

\newcommand{\aggrAttr}{aggregation attribute\xspace}
\newcommand{\aggrAttrs}{aggregation attributes\xspace}

\begin{document}
\begin{frontmatter}
\title{Object-centric Process Predictive Analytics}

\author[1,2]{Riccardo Galanti\corref{cor1}}
\ead{riccardo.galanti@ibm.com}

\author[2]{Massimiliano de Leoni\corref{cor1}}
\ead{deleoni@math.unipd.it}

\author[2]{Nicolò Navarin\corref{cor1}}
\ead{nnavarin@math.unipd.it}

\author[1]{Alan Marazzi\corref{cor1}}
\ead{alan.marazzi@ibm.com}

\address[1]{IBM, Bologna, Italy}
\address[2]{University of Padua, Padua, Italy}

\begin{abstract}
Object-centric processes (a.k.a. Artifact-centric processes) are implementations of a paradigm where an instance of one process is not executed in isolation but interacts with other instances of the same or other processes. Interactions take place through bridging events where instances exchange data. Object-centric processes are recently gaining popularity in academia and industry, because their nature is observed in many application scenarios. This poses significant challenges in predictive analytics due to the complex intricacy of the process instances that relate to each other via many-to-many associations. Existing research is unable to directly exploit the benefits of these interactions, thus limiting the prediction quality. This paper proposes an approach to incorporate the information about the object interactions into the predictive models. The approach is assessed on real-life object-centric process event data, using different KPIs. The results are compared with a na\"ive approach
that overlooks the object interactions, thus illustrating the benefits of their use on the prediction quality.
\end{abstract}


\end{frontmatter}

\section{Introduction}

Traditionally, processes are seen as being instantiated in cases that are constituted by  single flows that are executed in isolation. However, inter-organization processes are oftentimes more complex: usually, several instances of different processes are being executed at the same time and they may interact with each other.
In fact, the situation is more similar to choreographies where one instance of a process $P_1$ interacts and synchronizes with several instances of a second process $P_2$, and the other way around: one instance of $P_2$ might synchronize with multiple instances of $P_1$. The situation can be even more complex: instances of $P_2$ may in turn interact with instances of some $P_3$, and so on.
For instance, consider a retail shop in Padua (Italy): several customers may order products manufactured in a factory in Brisbane (Australia). The factory associates many customer orders to a single manufacturer order to save money. Also, the same customer orders can include products from different manufacturers in different parts of the globe. Customer's and manufacturer's orders are managed via instances of different processes: one instance of customer-order process can be associated to several of manufacturer-order process, and the other way round: each manufacturer-order process instance may be associated to many consumer-order ones.

The paradigm of object-centric processes (a.k.a. artifact-centric processes) is gaining more and more momentum in the recent years to model inter-organizational processes more naturally~\cite{10.1145/3297280.3297287,10.1007/978-3-030-30446-1_1,DBLP:conf/dlog/AalstAMT17}.
Any process execution materializes itself as a set of instances of the same/different processes that represent the life cycles of different objects (a.k.a.\ artifacts) that contribute to the process execution (e.g., the order and the delivery object). These processes for the different objects run independently and synchronize through some bridging events to exchange data needed to progress further.
The importance of object-centric process approaches is particularly evident in the domain of Process Mining: the IEEE Task Force has recently reported on the results of a survey with academics, practitioners, consultants and vendors in Process Mining, according to which only 33\% of the respondents consider forcing to pick one single case identifier to be a minor problem~\cite{XES-Survey}.

This paper focuses on object-centric process predictive analytics. This is a challenging problem that cannot be tackled via the current research in process predictive analytics, since the latter relies on the assumption that the instances refer to a single process.
In a nutshell, the main idea is that a main object type (e.g., \emph{Customer Order}) is chosen as viewpoint. The complex object interaction is unfolded in traces around the viewpoint:  one trace is created per object $o$ of the viewpoint type, and includes the events related to that object (i.e., process instance), and related to the objects of the same or different types that synchronize with $o$, directly or indirectly. When the complex interaction is unfolded in a multiset of traces, we can specialize the current state of the art in predictive analytics.

Process stakeholders define the Key Performance Indicators (KPIs) using domain knowledge: this corresponds to instantiating a  function $\MT(\sigma)$ that returns the KPI value for any trace $\sigma$. Since the viewpoint determines how events are grouped in traces, the determination of the appropriate KPI influences which viewpoint to choose. As an example, if the KPI relates to the duration of an order process, the viewpoint needs to be the order object.

The proposed approach is assessed through experiments on a real event log extracted from an object-centric process executed by an utility provider company in Italy. Experiments were conducted on different KPIs, illustrating that the complex interactions of  object-centric  processes  need to  be  taken  into  account  when  predicting, as this allows to consistently improve the predictive performances over simpler techniques.

Section~\ref{sec:Preliminary} introduces the preliminary concepts that are used later on: object-oriented and single-id event logs, KPIs, and the traditional problem of process predictive analytics. Section~\ref{sec:framework description} illustrates the procedure to create single-id event logs from object-oriented event logs, also illustrating how \aggrAttrs can be additionally used to better capture the object-to-object synchronization dynamics.
Section~\ref{sec:experiments} sketches some implementation details and extensively reports on the experimental phase, while related works are discussed in Section~\ref{sec:related}.
Finally, Section~\ref{sec:conclusions} concludes the paper, summarizing the paper context and contribution, and delineating some potential avenues of future work.

\section{Preliminaries}
\label{sec:Preliminary}

\subsection{Object-Centric Event Logs}

Object-centric processes are carried on with the support of one or more information systems. It is possible to extract the history of past executions into a transactional data set organized in form of object-centric event logs~\cite{OCEL}:
\begin{definition}[Object-Centric Event Log]
\label{def:o-clog}
Let $T$ be the universe of the timestamps.
An object-centric event log is a tuple $L= (E, A, AN, AV, AT, OT, O, \pi_{typ}, \pi_{act}, \pi_{time},\linebreak \pi_{vmap}, \pi_{omap}, \pi_{otyp}, \pi_{ovmap},  <)$ such that:
\begin{itemize}
\item $E$ is the set of event identifiers,
\item $A$ is the set of activity names,
\item $AN$ is  the  set  of  attributes  names,
\item $AV$ is the set of attribute values (with the requirement that $AN \cap AV =\emptyset$),
\item $AT$ is the set of attribute types,
\item $OT$ is the set of object types,
\item $O$ is the set of object identifiers,
\item $\pi_{typ}: AN \cup AV \rightarrow AT$ is  the  function  associating  an  attribute  name  or  value to  its  corresponding  type,
\item $\pi_{act}: E \rightarrow A$ is the function associating an event identifier to its activity,
\item $\pi_{time}: E \rightarrow T$ is the function associating timestamps to event identifiers,
\item $\pi_{vmap}: E \rightarrow (AN \not\rightarrow AV)$ is the function associating every event identifier $e \in E$ to a variable-to-value assignment function $val$ such that, for each attribute $a \in AN$ in the domain of $val$, $val(a)$ indicates the value assigned to $a$ by $e$,\footnote{The notation $\not\rightarrow$ indicates a partial function.}
\item $\pi_{omap}: E \rightarrow 2^O$ is  the  function  associating  an  event identifier  to  a  set  of related  object  identifiers,
\item $\pi_{otyp}: O \rightarrow OT$ assigns  precisely  one  object  type  to  each  object  identifier,
\item $<\; \subseteq (E \times E)$ is a partial order of events.\footnote{Typically, the partial order is induced by the timestamp,  i.e., $e' < e'' \iff \pi_{time}(e') < \pi_{time}(e'') $. However, we do not require to make that assumption.}
\end{itemize}
\end{definition}
\begin{example}
Table \ref{tab:object-centric_event_log_example} shows an excerpt of an object-centric event log of an Italian utility provider company.
It consists of five object types, each with its own object identifier: \texttt{Contract}, \texttt{Requisition}, \texttt{Order}, \texttt{Receipt}, \texttt{Invoice}. The first is \texttt{Contract}, which is the process concerning the stipulation of a contract with a customer, possibly followed by a \texttt{Requisition}, which is an optional process executed when the order needs a purchase requisition.
The \texttt{Order} process consists of several activities representing mainly quantity, price, or date modifications of the order, eventually approved by the Head of the department.
The \texttt{Receipt} process is then related to the receiving of the goods or the services requested, followed by the \texttt{Invoice} process, which includes everything related to payments. Some events are associated to a single object identifier, others have multiple (i.e., the so-called bridge events) that enable the synchronization and data exchange between objects. Figure \ref{fig:Uml_diagram} illustrates how objects are related to each other for synchronization and data exchanges. Note that relationships can be of many-to-many or many-to-one nature.
Events are associated with attributes, features  based on the properties of requisitions, orders (e.g., \texttt{order\_price}), receipts (e.g., \texttt{receipt\_quantity}), and invoices.

\begin{table*}[t!]
\caption{Example of object-centric event log. Each row represents an event. The blank spaces represent attributes missing values.}
\footnotesize
\centering
\resizebox{\textwidth}{!}
{\begin{tabular}{lllllllllllll}
\hline
{id} &
{\begin{tabular}[c]{@{}c@{}}activity\end{tabular}} &
{\begin{tabular}[c]{@{}c@{}}timestamp\end{tabular}} &
{\begin{tabular}[c]{@{}c@{}}Contract\end{tabular}} &
{\begin{tabular}[c]{@{}c@{}}Requisition\end{tabular}} &
{\begin{tabular}[c]{@{}c@{}}Order\end{tabular}} &
{\begin{tabular}[c]{@{}c@{}}Receipt\end{tabular}} &
{\begin{tabular}[c]{@{}c@{}}Invoice\end{tabular}} &
{\begin{tabular}[c]{@{}c@{}}user\end{tabular}} &
{\begin{tabular}[c]{@{}c@{}}order\_price\end{tabular}} &
{\begin{tabular}[c]{@{}c@{}}order\_delivery\_month\end{tabular}} &
{\begin{tabular}[c]{@{}c@{}}order\_purch\_group\end{tabular}} &
{\begin{tabular}[c]{@{}c@{}}rec\_quantity\end{tabular}} \\
\hline
e1 & Contract Line Creation & 2017-07-11 9:00 & c1 & & & & & CO01 & & & \\

e2 & Purch Contract Item Material Group Changed & 2017-07-14 11:00 & c1 & & & & & CO01 & & & & \\

e3 & Purchase Requisition Line Created & 2017-07-15 12:00 & c1 & rq1 & & & & A456 & & & & \\

e4 & Purchase Requisition Line Created & 2017-07-15 15:00 & c1 & rq2 & & & & A457 & & & & \\

e5 & Purchase Order Line Creation & 2017-07-16 15:00 & c1 & & o1 & & &
 A458 & 100 & 7 & 100\_L50 & \\

e6 & Purchase Order Line Creation & 2017-07-17 15:00 & & rq1 & o2 & & &
 A458 & 200 & 8 & 100\_L51 & \\

e7 & Purchase Order Line Creation & 2017-07-18 15:00 & & rq2 & o3 & & &
 A458 & 300 & 8 & 100\_L52 & \\

e8 & Goods Line Registered & 2017-07-22 15:00 & & & o1 & r1 & & A456 &
 100 & 7 & 100\_L50 & 10 \\

e9 & Invoice Receipt & 2017-07-22 16:00 & & & & & i1 & A125 & & & & \\

e10 & Purchase Requisition Group Changed & 2017-07-22 19:00 & & rq1 & & & & A456 & & & & \\

e11 & Purchase Order Line Creation & 2017-07-23 9:00 & & rq1 & o4 & & &
 A458 & 600 & 8 & 100\_L51 & \\

e12 & Goods Line Registered & 2017-07-23 15:00 & & & o2 & r2 & & A456 &
100 & 8 & 100\_L50 & 10 \\

e13 & Invoice Registered & 2017-07-29 11:00 & & & & r1,r2 & i1 & A125 &
 & & & 10 \\

e14 & Invoice Cleared & 2017-07-30 12:00 & & & & & i1 & A125 & & & & \\

e15 & Goods Line Registered & 2017-07-31 15:00 & & & o4 & r3 & & A456 &
600 & 8 & 100\_L51 & 10 \\

e16 & Invoice Registered & 2017-08-10 11:00 & & & & r2,r3 & i2 & A125 &
& & & 10 \\

e17 & Invoice Cleared & 2017-08-15 14:00 & & & & & i2 & A125 & & & & \\

e18 & Goods Line Registered & 2017-08-16 15:00 & & & o3 & r4 & & A456 & 300
& 8 & 100\_L52 & 5 \\

e19 & Purchase Requisition Supplier Changed & 2017-08-16 17:00 & & rq2 & & & & A456 & & & & \\

e20 & Invoice Registered & 2017-08-18 11:00 & & & & r4 & i3 & A125 &
& & & 5 \\

e21 & Invoice Cleared & 2017-08-20 14:00 & & & & & i3 & A125 & & & & \\

\hline

\end{tabular}}
\label{tab:object-centric_event_log_example}
\end{table*}

\begin{figure}[ht!]
    \centering
    \includegraphics[width=\textwidth]{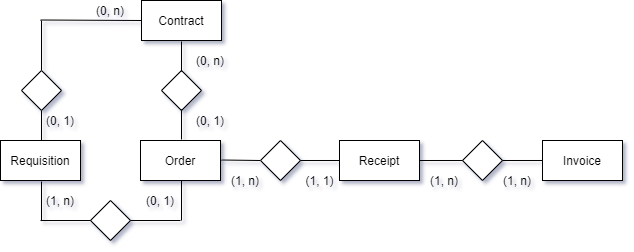}
    \caption{Diagram representing cardinality between the different object types in the considered object-centric event log. For each object type, the cardinality with the subsequent or the previous object type is represented as (\textit{min\_cardinality}, \textit{max\_cardinality})}
    \label{fig:Uml_diagram}
\end{figure}
\end{example}

\subsection{Single-Id Process Predictive Analytics}
\label{sec:LiteraturePrediction}

In this section, we discuss the typical techniques adopted to train a predictive model starting from an event log with a single identifier.
\begin{definition}[Single-id Event Log]\label{def:mp-simplog}
\noindent A single-id event log is a tuple: $\ML=(E,T,A,$ $AN,AV,AT, \pi_{typ}, \pi_{act}, \pi_{time}, \pi_{vmap})$ consisting of a set $E$ of event identifiers, a set $T \subset E^*$ of traces, i.e. sequences of event identifiers, a set $A$ of activity names, sets $AN$ and $AV$ of attribute names and values, a function $\pi_{typ}$ associating
attribute names and values to types, $\pi_{act}: E \rightarrow A, \pi_{time}: E \rightarrow T, \pi_{vmap}: E \rightarrow (AN \not\rightarrow AV)$\footnote{Notation $A \not\rightarrow B$ is used to highlight a partial function from some elements of a domain $A$ to elements of $B$} associating each event identifier to event's activity, timestamp, and attribute assignment, respectively.
\end{definition}
The aim is to predict the value of a key performance indicator (KPI), which depends on the specific process domain. This corresponds to providing the definition a KPI function.
\begin{definition}[KPI Function]
\label{def:KPI}
Let $\ML=(E,T,A,AN,AV,AT, \pi_{typ}, \pi_{act}, \pi_{time}, \linebreak\pi_{vmap})$ be a single-id event log. Let $\MW_K$ be the set of possible KPI values. A KPI is a function $\MT_\ML: E^* \times \NN \not\rightarrow \MW_K$ such that, given a trace $\sigma \in E^*$ and an integer index $i \leq |\sigma|$, $\MT_\ML(\sigma,i)$ returns the KPI value of $\sigma$ after the occurrence of the first $i$ events.\footnote{Given a sequence $X$, $|X|$ indicates the length of $X$.}
\end{definition}
Note that our KPI definition assumes it to be computed a posteriori, when the execution is completed and leaves a complete trail as a certain trace $\sigma$. In many cases, the KPI value is updated after each activity execution, which is recorded as next event in trace; however, other times, this is only known after the completion. In the remainder, when clear from the context, we often omit the subscript $\ML$. 

We aim to be generic and account for all relevant domains. Given a trace  $\sigma = \langle e_1,\ldots,e_n \rangle$ that records a complete process execution, the following are example of three potential KPI definitions:
\begin{description}
\item [Remaining Time.]  This corresponds to the situation in which, after executing a sequence of $i$ events, the KPI measures how long the process execution will still last to completion: $\MT_{remaining}(\sigma,i)=\pi_{time}(e_n)-\pi_{time}(e_i)$, namely the difference between the timestamp of latest future trace event $e_n$ and that of the last occurred event $e_i$.
\item [Activity Occurrence.] It measures whether a certain activity is going to eventually occur in the future, such as an activity \emph{Open Loan} in a loan-application process. The corresponding KPI definition for the occurrence of an activity $A$ is $\MT_{occur\_A}(\sigma,i)$, which is equal to true if activity $A$ occurs in $\langle e_{i+1},\ldots,e_n \rangle$ and $i<n$; otherwise false.
\item [Customer Satisfaction.] This is a typical KPI for several service providers. Let us assume, without losing generality, to have a trace $\sigma = \langle e_1,\ldots,e_n \rangle$ where the satisfaction is known at the end, e.g.\ through a questionnaire. Assuming the satisfaction level is recorded with the last event - say $e_n(sat)$ . Then, $\MT_{cust\_satisf}(\sigma,i)=e_n(sat)$.
\end{description}

We can define the prediction problem on unfolded logs as follows.
\begin{definition}[Prediction Problem on Single-id Event Logs]
Let $\ML$ be a single-id event log that records the execution of a given process, for which a KPI $\MT_\ML$ is defined. Let $\sigma = \langle e_1,\ldots,e_k \rangle$ be the  trace of a running case, which eventually will complete as \linebreak \mbox{$\sigma_T=\langle e_1,\ldots,e_k, e_{k+1}\ldots,e_n \rangle$}.
The prediction problem can be formulated as forecasting the value of $\MT_\ML(\sigma_T,i)$ for all $k < i \leq n$.
\end{definition}
In the process mining literature, this problem has been faced with different machine learning models~\cite{Marquez-Chamorro18,Park19,TaxVRD17,LSTM_time,Polato_Sperduti_Burattin_Leoni_2018,survey_remaining_time_verenich}.

The training set is composed by pairs $(x,y) \in \mathcal{X} \times \mathcal{Y}$ where $\mathcal{X}$ encodes the independent variables (also known as \textbf{features}) with their values, and $\mathcal{Y}$ is the the dependent variable with its value (i.e.\ the value to predict).
Process predictive analytics requires a KPI definition $\overline{\MT}$ as input (cf.\ Definition~\ref{def:KPI}). Let $\MW_K={img}(\overline{\MT})$ be the domain of possible KPI values (i.e.\ the image/codomain of $\overline{\MT}$: $\mathcal{Y}=\MW_K$.
Afterwards, each prediction technique requires the definition of the domain $\mathcal{X}$ and a \textbf{trace-to-instance encoding function} $\rho: \ME^* \rightarrow \mathcal{X}$, which maps each (prefix of a) trace $\sigma$ in an element $\rho(\sigma)\in\mathcal{X}$.

The prediction model is trained off-line via a dataset $\mathcal{D}$ that is created from an event log $\ML$ as follows. Each prefix $\sigma$ of each each trace $\sigma_T \in \ML$ generates one distinct item in $\mathcal{D}$ consisting of a pair $(x,y) \in (\mathcal{X} \times \mathcal{Y})$
 where $x=\rho(\sigma)$ and $y=\overline{\MT}(\sigma_T,|\sigma|)$.
Once the dataset item of every trace prefix is created, the model is trained. The resulting prediction model (a.k.a.\ predictor) can be abstracted as an oracle function $\Phi_\mathcal{D}:\mathcal{X} \rightarrow \mathcal{Y}$.

This paper does not focus on comparing different machine learning techniques. Instead, we focus on how it is possible to leverage on methodologies developed for single-process prediction problems to tackle the object-centric process prediction problem. In the implementation and experiments, we use the Catboost method~\cite{Catboost}, a high-performance open source framework for gradient boosting on decision trees, but different types of predictive models could be learnt. The choice fell onto Catboost because experiments show that
it outperforms and solves limitations of current state-of-the-art implementations of gradient boosted decision trees~\cite{Catboost}.
It is backed by solid theoretical results that explain how strong predictors can be built by iteratively
combining weaker models (base predictors) in a greedy manner.
Catboost, in particular, at each iteration $t$ of the algorithm performs a random permutation of the features and a tree is constructed on the basis of it.
Moreover, for each split of a tree, CatBoost combines (concatenates) all categorical features (and their combinations) already used for previous splits in the current tree with all categorical features in the dataset.

In the domain of Catboost learning, the definition of the trace-to-instance encoding function requires the intermediate concept of \textit{event-to-tuple function} \linebreak $\zeta_\ML: \ME \rightarrow A \times AT \times (AV)^w$, which encodes each event of a single-id event log \linebreak $\ML=(E,T,A,AN,AV,AT, \pi_{typ}, \pi_{act}, \pi_{time}, \pi_{vmap})$, where $w=|AN|$ is the number of attributes defined in the event log $\ML$.
The event-to-tuple function is defined as follows.
In particular, indicated the concatenation of two tuples with $\oplus$,
given an event in $\ML$ with identifier $e_i$,
$\zeta_\ML(e)=[\pi_{act}(e),\pi_{time}(e)] \displaystyle\bigoplus_{v \in AN} \pi_{vmap}(v)$.\footnote{To keep the explanation simple, we assume that the enumerations of all variables $v$ in $AN$ are always returned consistently, as if there is a total order among the variables (e.g., the alphabetical order).}

In Catboost, the trace-to-instance encoding function also considers the history of each partial trace $\sigma$ by considering the number of times that each activity has been performed in $\sigma$. Consequently, we define the function $\rho^{aggr}_\ML(\langle e_1, \ldots, e_m \rangle)$; here, for each activity $a \in A$, one dimension exists in $\rho^{aggr}_\ML(\sigma): \ME^* \rightarrow (\NN)^{|A|}$ that takes on a value equal to the number of events $\overline{e} \sigma$ that refer to $a$, i.e.\ such that $\pi_{act}{\overline{e}}$.
The function $\rho_\ML$ is then defined as: $\rho_\ML(\langle e_1, \ldots, e_m \rangle)=\rho^{aggr}_\ML(\langle e_1, \ldots, e_m \rangle) \bigoplus \zeta_\ML(e_m)$.

\subsection{Explanations of Process Predictions}
\label{sec:explanations}
Several prediction models, including Catboost, are black boxes, making it difficult to explain the predictions, namely to determine the degree with which each feature influences the predictions. Explaining the predictions is beneficial to build user trust in the prediction model. The remainder briefly summarizes the basic concepts behind the framework for prediction explanation that has been introduced in~\cite{galanti2020explainable}.
This framework has been used during the evaluation to illustrate that several features related to the object interaction have a significant impact on the predictions, thus consequently showing their relevance to improve the prediction accuracy.

The  framework in~\cite{galanti2020explainable} leverages on the Shapley Values~\cite{shapley1953value}, which is a game theory approach to fairly distribute the payout among the players that have collaborated in a cooperative game.
The assumption is that the features from an instance correspond to the players, and the payout is the difference between the prediction made by the predictive model and the average prediction (also called {\em base value}).
Intuitively, given a predicted instance, the Shapley Value  of a feature expresses how much the feature value contributes to the model prediction~\cite{molnar2019}:
\begin{definition}[Shapley Value]
Let $X=\{x_1,\ldots,x_n\}$ be a set of features. The Shapley value for feature $x_i$ is defined as:
\begin{equation*}
\resizebox{\columnwidth}{!}{$
   \psi_i=\sum_{S\subseteq\{ x_1, \ldots, x_m \}\setminus\{x_i\}}\frac{|S|!\left(p-|S|-1\right)!}{p!}\left(val\left(S\cup\{x_i\}\right)-val(S)\right)
   $}
\end{equation*}
where $val(X')$ is the so-called payout for only using the set of feature values in $X' \subset X$ in making the prediction.
 \label{def:SV}
\end{definition}
Intuitively, the formula in Definition~\ref{def:SV} evaluates the effect of incorporating the feature value $x_i$ into any possible subset of the feature values considered for prediction. In the equation, variable $S$ runs over all possible subsets of feature values, the term $val\left(S\cup\{x_i\}\right)-val(S)$
corresponds to the marginal value of adding $x_i$ in the prediction using only the set of feature values in $S$,
and the term
$\frac{|S|!\left(p-|S|-1\right)!}{p!}$ corresponds to all the possible permutations with subset size $|S|$, to weight different sets differently in the formula.
This way, all possible subsets of attributes are considered, and the corresponding effect is used to compute the Shapley Value of $x_i$.

The starting point for the explainable framework is the trace-to-instance encoding function $\rho: \ME^* \rightarrow \mathcal{X}$ (cf.\ Section~\ref{sec:LiteraturePrediction}), and a single-id event log $\ML=(E,T,A,AN,AV,$  $AT, \pi_{typ}, \pi_{act}, \pi_{time}, \pi_{vmap})$.

Let us recall that given a trace $\sigma=\langle e_1, \ldots, e_m \rangle \in T$ , $\rho(\sigma)=[x^1, \ldots,x^n]$, each feature $f^i$ has an associated value $x^i$. As mentioned in Section~\ref{sec:LiteraturePrediction}, a feature $f^i$ can be of different nature, such as a process attribute, a timestamp, or the number of executions of an activity in $\sigma$. The prediction model is built over the multiset $\uplus_{\sigma' \in T} \rho(\sigma')$.

When applied for prediction explanations, the Shapley values for a trace $\sigma$ are computed over tuple $\rho(\sigma)=[x^1, \ldots,x^n]$, thus resulting in a tuple of Shapley values $\Psi=[\psi^1, \ldots ,\psi^n]$, with $\psi^i$ being the Shapley value of feature $f^i$. In accordance with the Shapley values theory, the explanation of $\psi^i$ is as follows: since feature $f^i=x^i$, the KPI prediction deviates $\psi^i$ units from the average KPI value of the event-log traces.

The computation is the Shapley value is repeated for each trace of $\ML$. However, if $f^i$ is numerical, several different values can be observed for $f^i$, yielding
a large number of explanations $f^i=x^i_1, \ldots f^i=x^i_k$. Some of these explanations are equivalent from a domain viewpoint: e.g., $amount=10000$, $amount=10050$ might be referring to the same class of amount in a loan application. Therefore, $q$ representative values $w^i_1, \ldots, w^i_q$ are selected out of values $x^i_1, \ldots x^i_k$ (namely, with $q \ll k$) so as to obtain explanations of type $f^i<w^i_1$, $w^i_1 \leq f^i <w^i_2$, $\ldots$, $f^i \geq w^i_q$. Values $w^i_1, \ldots, w^i_q$ can be obtained taking the boundaries of the buckets obtained via discretization techniques. In particular, our implementation operationalizes a discretization of each feature $f^i$ on the basis of decision/regression as follows. The training set consists of tuple with only two features: $f^i$ used as the independent variable, and  the KPI as target/dependent variable. The values observed at the splits of the tree nodes induce the boundaries and, consequently, the buckets.

While an exact computation of the Shapley values requires to consider all combinations of features, efficient estimations can be obtained through polynomial algorithms that use greedy approaches~\cite{molnar2019}.

The discussion so far focused on the Shapley values for single (prefixes of) traces. It is possible to show the distribution over multiple traces via, e.g., boxplots. Examples will be shown in Section \ref{sec:experiments} (cf. Figures \ref{inv_rec_aggr_boxplot}, \ref{inv_cl_no_aggr_boxplot} and \ref{inv_cl_aggr_boxplot}).

\section{Predictive Analytics in Object-Centric Processes}
\label{sec:framework description}

The starting point is an object-centric log $L= (E, A, AN, AV, AT, OT, O, \pi_{typ},\linebreak \pi_{act},\pi_{time}, \pi_{vmap}, \pi_{omap}, $ $\pi_{otyp}, \pi_{ovmap},  <)$. Our object-centric process prediction requires analysts to decide a so-called \textit{viewpoint}, which is an object type  $o_t \in OT$ of the process (e.g., \textit{Requisition}). This defines how the events in an object-centric event log are aggregated to form traces of a single-id event log.

A single-id event log $\ML=(E',T,A',AN',AV',AT', \pi'_{typ}, \pi'_{act}, \pi'_{time}, \pi'_{vmap})$ is created from a chosen viewpoint $o_t \in OT$ as follows.
The trace set $T$ contains one trace $\sigma_o$ for each object $o \in O$ such that $\pi_{otyp}(o)=o_t$. To determine which events to include in $\sigma_o$, we compute the timestamp of the first event in $E$ that has $o$ as one of the object identifiers: 
\begin{equation}
    t_{o} = \underset{e \in E.\; o \in \pi_{omap}(e)}{min}\; {\pi_{time}(e)}     
\end{equation}
Trace $\sigma_o$ will include every event $e \in E$ with timestamp larger  than or equal to $t_o$ such that it at least contains $o$ as identifier (namely, such that $\{o\} \subseteq \pi_{omap}(e)$) or contains an identifier of an object $o'$ in a certain set $R^+_L(o)$ of related objects (namely such that $R^+_L(o) \cap \pi_{omap}(e) \neq \emptyset$). The constraint to exclude events with timestamp smaller than $t_o$ is motivated by the fact that, if an event $\overline{e}$ precedes the first event $e_1$ of $\sigma_o$, $\overline{e}$ is unrelated to the behavior in $\sigma_o$: if $\overline{e}$ had influenced or was influenced by the behavior in $\sigma_o$, $\overline{e}$ would have occurred after $e_1$, which is, in fact, intended as the creation event of the process related to life cycle of $o$. 

The set $R^+_L(o)$ is constructed as follows. Let us define $R^1_{L,o_t}(o)$ as the set of objects directly related to $o$ of type different than $o_t$:
 \[
R^1_{L,o_t}(o) = \{ o' \in O : \exists e \in E.\; \{o,o' \} \subseteq \pi_{omap}(e) \land \pi_{otyp}(o') \neq o_t \}.
\]
We can also define the set of objects of type different than $o_t$ that are indirectly related $o$ via a ``bridge'' object $o'$ of type  different than $o_t$:
\begin{equation}
R^2_{L,o_t}(o) = \underset{o' \in R^{1}_{L,o_t}(o)}{\bigcup}  R^1_{L,o_t}(o')
\end{equation}
The definition above can also be extended for any $R^i_{L,o_t}(o)$ with $i>1$ as follows: 
\begin{equation}
R^i_{L,o_t}(o) = \underset{o' \in R^{i-1}_{L,o_t}(o)}{\bigcup}  R^1_{L,o_t}(o')
\end{equation}
The set $R^+_L(o)$ can thus be defined as the union, for all integer indexes $i \geq 1$, of the sets of the  objects of type different than $\pi_{otyp}(o)$ that are related to $o$ via $(i-1)$ briding events of type different than $\pi_{otyp}(o)$:
\begin{equation}
R^+_L(o) = {\bigcup_{i=1}^\infty R^i_{L,\pi_{otyp}(o)}(o)}
\end{equation}
The application of the aforementioned procedure to each object $o$ of the viewpoint type creates the set $T$ of traces, which in turn induces the the other elements of the $\ML$ tuple as follows. The set $E' = \cup_{\sigma_o \in T} \cup{e' \in \sigma_o} e$ of event identifier contains the event identifiers in $T$. The domain of attribute names, values and types is the same as in the object-centric log: $AN'=AN$, $AV'=AV$, $AT'=AT$, and $\pi'_{typ}=\pi_{typ}$. Functions $\pi'_{act}, \pi'_{time}, \pi'_{vmap}$ are restricted over domain $E'$, namely for each $e' \in E'$ $\pi'_{act}(e)=\pi_{act}(e)$, $\pi'_{time}(e)=\pi_{time}(e)$, $\pi'_{vmap}(e)=\pi_{vmap}(e)$.

\begin{table*}[t!]
\caption{Unfolded event log obtained from the object-centric event log in Table \ref{tab:object-centric_event_log_example} with the proposed object-centric approach when considering the Requisition viewpoint. The horizontal lines split the different traces. The blank spaces represent attributes missing values.}
\footnotesize
\centering
\resizebox{\textwidth}{!}
{\begin{tabular}{lllllllllllll}
\hline
{id} &
{\begin{tabular}[c]{@{}c@{}}activity\end{tabular}} &
{\begin{tabular}[c]{@{}c@{}}timestamp\end{tabular}} &
{\begin{tabular}[c]{@{}c@{}}Contract\end{tabular}} &
{\begin{tabular}[c]{@{}c@{}}Requisition\end{tabular}} &
{\begin{tabular}[c]{@{}c@{}}Order\end{tabular}} &
{\begin{tabular}[c]{@{}c@{}}Receipt\end{tabular}} &
{\begin{tabular}[c]{@{}c@{}}Invoice\end{tabular}} &
{\begin{tabular}[c]{@{}c@{}}user\end{tabular}} &
{\begin{tabular}[c]{@{}c@{}}order\_price\end{tabular}} &
{\begin{tabular}[c]{@{}c@{}}order\_delivery\_month\end{tabular}} &
{\begin{tabular}[c]{@{}c@{}}order\_purch\_group\end{tabular}} &
{\begin{tabular}[c]{@{}c@{}}rec\_quantity\end{tabular}}\\
\hline
\hline

e3 & Purchase Requisition Line Created & 2017-07-15 12:00 & c1 & rq1 & & & & A456 & & & & \\

e4 & Purchase Requisition Line Created & 2017-07-15 15:00 & c1 & rq2 & & & & A457 & & & & \\

e5 & Purchase Order Line Creation & 2017-07-16 15:00 & c1 & & o1 & & &
 A458 & 100 & 7 & 100\_L50 & \\

e6 & Purchase Order Line Creation & 2017-07-17 15:00 & & rq1 & o2 & & &
 A458 & 200 & 8 & 100\_L51 & \\

e7 & Goods Line Registered & 2017-07-22 15:00 & & & o1 & r1 & & A456 &
 100 & 7 & 100\_L50 & 10 \\

e9 & Invoice Receipt & 2017-07-22 16:00 & & & & & i1 & A125 & & & & \\

e10 & Purchase Requisition Group Changed & 2017-07-22 19:00 & & rq1 & & & & A456 & & & & \\

e11 & Purchase Order Line Creation & 2017-07-23 9:00 & & rq1 & o4 & & &
 A458 & 600 & 8 & 100\_L51 & \\

e12 & Goods Line Registered & 2017-07-23 15:00 & & & o2 & r2 & & A456 &
200 & 8 & 100\_L50 & 10 \\

e13 & Invoice Registered & 2017-07-29 11:00 & & & & r1,r2 & i1 & A125 &
 & & & 10 \\

e14 & Invoice Cleared & 2017-07-30 12:00 & & & & & i1 & A125 & & & & \\

e15 & Goods Line Registered & 2017-07-31 15:00 & & & o4 & r3 & & A456 &
600 & 8 & 100\_L51 & 10 \\

e16 & Invoice Registered & 2017-08-10 11:00 & & & & r2,r3 & i2 & A125 &
& & & 10 \\

e17 & Invoice Cleared & 2017-08-15 14:00 & & & & & i2 & A125 & & & & \\

\hline

e4 & Purchase Requisition Line Created & 2017-07-15 15:00 & c1 & rq2 & & & & A457 & & & & \\

e7 & Purchase Order Line Creation & 2017-07-18 15:00 & & rq2 & o3 & & &
 A458 & 300 & 8 & 100\_L52 & \\

e18 & Goods Line Registered & 2017-08-16 15:00 & & & o3 & r4 & & A456 & 300
& 8 & 100\_L52 & 5 \\

e19 & Purchase Requisition Supplier Changed & 2017-08-16 17:00 & & rq2 & & & & A456 & & & & \\

e20 & Invoice Registered & 2017-08-18 11:00 & & & & r4 & i3 & A125 &
& & & 5 \\

e21 & Invoice Cleared & 2017-08-20 14:00 & & & & & i3 & A125 & & & & \\
\hline

\end{tabular}}
\label{tab:event_log_our_approach_req}
\end{table*}

\begin{table*}[t!]
\caption{Unfolded event log obtained when considering the Requisition viewpoint from the object-centric event log in Table \ref{tab:object-centric_event_log_example} with the existing approach, assuming each instance belongs to a single process. The horizontal lines split the different traces. The blank spaces represent attributes missing values.}
\footnotesize
\centering
\resizebox{\textwidth}{!}
{\begin{tabular}{lllllllllllll}
\hline
{id} &
{\begin{tabular}[c]{@{}c@{}}activity\end{tabular}} &
{\begin{tabular}[c]{@{}c@{}}timestamp\end{tabular}} &
{\begin{tabular}[c]{@{}c@{}}Contract\end{tabular}} &
{\begin{tabular}[c]{@{}c@{}}Requisition\end{tabular}} &
{\begin{tabular}[c]{@{}c@{}}Order\end{tabular}} &
{\begin{tabular}[c]{@{}c@{}}Receipt\end{tabular}} &
{\begin{tabular}[c]{@{}c@{}}Invoice\end{tabular}} &
{\begin{tabular}[c]{@{}c@{}}user\end{tabular}} &
{\begin{tabular}[c]{@{}c@{}}order\_price\end{tabular}} &
{\begin{tabular}[c]{@{}c@{}}order\_delivery\_month\end{tabular}} &
{\begin{tabular}[c]{@{}c@{}}order\_purch\_group\end{tabular}} &
{\begin{tabular}[c]{@{}c@{}}rec\_quantity\end{tabular}} \\
\hline
\hline

e3 & Purchase Requisition Line Created & 2017-07-15 12:00 & c1 & rq1 & & & & A456 & & & & \\

e6 & Purchase Order Line Creation & 2017-07-17 15:00 & & rq1 & o2 & & &
 A458 & 200 & 8 & 100\_L51 & \\

e10 & Purchase Requisition Group Changed & 2017-07-22 19:00 & & rq1 & & & & A456 & & & & \\

e11 & Purchase Order Line Creation & 2017-07-23 9:00 & & rq1 & o4 & & &
 A458 & 600 & 8 & 100\_L51 & \\
\hline

e4 & Purchase Requisition Line Created & 2017-07-15 15:00 & c1 & rq2 & & & & A457 & & & & \\

e7 & Purchase Order Line Creation & 2017-07-18 15:00 & & rq2 & o3 & & &
 A458 & 300 & 8 & 100\_L52 & \\

e19 & Purchase Requisition Supplier Changed & 2017-08-16 17:00 & & rq2 & & & & A456 & & & & \\

\end{tabular}}
\label{tab:event_baseline_req}
\end{table*}

\begin{example}
Let us assume to unfold the object-oriented event log in Table \ref{tab:object-centric_event_log_example}, using \texttt{Requisition} as viewpoint. The result is presented in Table~\ref{tab:event_log_our_approach_req}.
Traces are split through horizontal lines. There are two traces, as many as the number of requisitions.
In particular, since in the first trace we focused on the requisition with identifier \texttt{rq1}, we first considered the events containing \texttt{rq1} as identifier (\texttt{e3}, \texttt{e6}, \texttt{e10}, \texttt{e11}); afterwards, we considered the events related transitively to \texttt{rq1}, i.e.\ with at least one identifier of an object in $R^+_L(rq1)$. For instance, we considered event \texttt{e5} because it contains the contract identifier \texttt{c1} that is related to \texttt{rq1} via event \texttt{e3}, which contains both \texttt{rq1} and \texttt{c1} among the identifiers.
Please note that, e.g., in the first trace we do not consider event \texttt{e19} because \texttt{e19} only contains the object identifier \texttt{rq2}, which is of the same type as \texttt{rq1}, namely the requisition object type. Neither do we consider such events as \texttt{e18} because, e.g., \texttt{e18} is only associated to \texttt{rq1} via a transitive relation that goes through \texttt{rq2}, which is of the same type as \texttt{rq1}. Event \texttt{e4} is instead included in the first trace because it is associated to the contract \texttt{c1}, which is associated transitively to the requisition \texttt{rq1}.
Event \texttt{e4} is also in the second trace because it is associated to \texttt{rq2}, since \texttt{rq2} is within the object identifiers of \texttt{e4}.
It is worth noting that events \texttt{e1} and \texttt{e2} are excluded from the first and second trace since their timestamps are smaller than \texttt{e3} or \texttt{e4}, which are respectively the first events with \texttt{rq1} or \texttt{rq2} as object identifiers.

Conversely, existing techniques would be unable to deal with multiple object types, namely with multiple interleaving processes; they would only restrict to one single process, which would likely be the process related to objects of type \texttt{Requisition}. Therefore, there would again be two traces, one per requisition. Differently from our approach, the event log would only contain the events that
present the first or the second requisition among the identifiers. The resulting log is in Table~\ref{tab:event_baseline_req}.
\label{example_viewpoint_events}
\end{example}

Summarising, the approach starts from an object-centric event log $L= (E, A, AN,\linebreak AV, $ $AT, OT, O, \pi_{typ}, \pi_{act},\pi_{time}, \pi_{vmap}, \pi_{omap}, $ $\pi_{otyp}, \pi_{ovmap},  <)$, which is unfolded to a single-id event log $\ML=(E',T,A',AN',AV',AT', \pi'_{typ}, \pi'_{act}, \pi'_{time}, \pi'_{vmap})$ around a viewpoint, namely an object type  $o_t \in OT$ of the process (e.g., \textit{Requisition}). The viewpoint defines how the events in an object-centric event log are aggregated to form traces of a single-id event log.

These traces can be used to train and test a proper prediction model using the off-the-shelf techniques discussed in Section~\ref{sec:LiteraturePrediction}. However, this unfolding to a single-id event log does not preserve information about the number of objects correlated to the viewpoint object. In fact, it excludes the information about the attributes associated with correlated objects, and their respective value. For instance, let us assume that we aim to use the total processing time of requisitions as KPI, and this time is somehow correlated to the number of orders associated to the requisitions (e.g., a requisition for more orders takes longer to be processed).
The prediction of this KPI could be more accurate if the number of orders objects associated with the requisition were used to train and use the prediction model. 

To mitigate this information loss, we extend our first approach to a richer one, where we introduce \aggrAttrs that synthesize additional interaction information. Given a prefix $\sigma'_o$, of the trace $\sigma_o$ for a viewpoint object $o$, the following \aggrAttrs are included.
\begin{enumerate}
\item A given attribute $a$ that can take different values $v_1,\ldots,v_n$ in a set $\{o'_1,\ldots,o'_m\} \subseteq R^+_L(o)$ of objects related to $o$. The standard encoding would only retain one value for $a$, namely the value observed in the latest event in $\sigma_o$ that assigns a value to  $a$.
We define:
\begin{itemize}
    \item if $a$ is numerical (i.e., $\pi_{typ}(a) \subseteq \RR$), one feature $f$ is added to the feature set that contains an aggregated value $\psi(v_1,\ldots,v_n)$ summarizing observed values. The aggregation function $\psi(\vec{v})$ may be customized depending on the domain, such as the average value in $\vec{v}$;
    
     \item if $a$ is categorical, a feature is added for each pair $(a,v_i)$ with value  defined as the ratio between the number of attribute assignments to the value $v_i$ over the total assignments of $a$.
    
\end{itemize}
\item For each object type $o'_t \in OT$, one feature encodes the number of objects of type $o'_t$ associated with events of $\sigma'_o$, namely
$|\{ o' \in O: \pi_{otyp}(o) = o'_t \land \exists e \in \sigma'_o.\; o' \in \pi_{omap}(e) \}|$.
\item For each object type $o'_t \in OT$ and for each activity $a \in A$
one feature is added with value equal to the percentage of objects of type $o'_t$ for which activity $a$ has occurred at least once in trace $\sigma'_o$.
\end{enumerate}

\begin{sidewaystable}
\caption{Unfolded event log obtained from the object-centric event log in Table \ref{tab:object-centric_event_log_example} with the proposed object-centric approach when considering the Requisition viewpoint and the aggregation attributes. The horizontal lines split the different traces. The blank spaces represent attributes missing values.}
\footnotesize
\centering
\resizebox{1.2\textwidth}{!}
{
\begin{tabular}{llllllllllllllllllllllllllllll}
\hline
{id} &
{\begin{tabular}[c]{@{}c@{}}activity\end{tabular}} &
{\begin{tabular}[c]{@{}c@{}}timestamp\end{tabular}} &
{\begin{tabular}[c]{@{}c@{}}Contract\end{tabular}} &
{\begin{tabular}[c]{@{}c@{}}Requisition\end{tabular}} &
{\begin{tabular}[c]{@{}c@{}}Order\end{tabular}} &
{\begin{tabular}[c]{@{}c@{}}Receipt\end{tabular}} &
{\begin{tabular}[c]{@{}c@{}}Invoice\end{tabular}} &
{\begin{tabular}[c]{@{}c@{}}user\end{tabular}} &
{\begin{tabular}[c]{@{}c@{}}order\_\\price\end{tabular}} &
{\begin{tabular}[c]{@{}c@{}}order\_\\delivery\_\\month\end{tabular}} &
{\begin{tabular}[c]{@{}c@{}}order\_\\purch\_\\group\end{tabular}} &
{\begin{tabular}[c]{@{}c@{}}rec\_\\quantity\end{tabular}} & {\begin{tabular}[c]{@{}c@{}}\# Orders\end{tabular}} & {\begin{tabular}[c]{@{}c@{}}\# Receipts\end{tabular}} & {\begin{tabular}[c]{@{}c@{}}\# Invoices\end{tabular}} & {\begin{tabular}[c]{@{}c@{}}Avg\\order\_\\price\end{tabular}} & {\begin{tabular}[c]{@{}c@{}}\% order\_\\delivery\_\\month=7\end{tabular}} & {\begin{tabular}[c]{@{}c@{}}\% order\_\\delivery\_\\month=8\end{tabular}} & {\begin{tabular}[c]{@{}c@{}}\% order\_\\purch\_\\group=\\100\_L50\end{tabular}} & {\begin{tabular}[c]{@{}c@{}}\% order\_\\purch\_\\group=\\100\_L51\end{tabular}} & {\begin{tabular}[c]{@{}c@{}}\% order\_\\purch\_\\group=\\100\_L52\end{tabular}} & {\begin{tabular}[c]{@{}c@{}}Avg\\rec\_\\quantity\end{tabular}} &
{\begin{tabular}[c]{@{}c@{}}(Order,\\\% Purchase\\Order Line\\Creation)\end{tabular}} &
{\begin{tabular}[c]{@{}c@{}}(Order,\\\% Goods\\Line\\Registered)\end{tabular}} &
{\begin{tabular}[c]{@{}c@{}}(Receipt,\\\% Goods\\Line\\Registered)\end{tabular}} &
{\begin{tabular}[c]{@{}c@{}}(Invoice,\\\% Invoice\\Receipt)\end{tabular}} &
{\begin{tabular}[c]{@{}c@{}}(Receipt,\\\% Invoice\\Registered)\end{tabular}} &
{\begin{tabular}[c]{@{}c@{}}(Invoice,\\\% Invoice\\Registered)\end{tabular}} &
{\begin{tabular}[c]{@{}c@{}}(Invoice,\\\% Invoice\\Cleared)\end{tabular}}\\
\hline
\hline

e3 & \makecell{Purchase Requisition\\Line Created} & \makecell{2017-07-15\\12:00} & c1 & rq1 & & & & A456 & & & & & 0 & 0 & 0 &  & 0 & 0 & 0 & 0 & 0 & & 0 & 0 & 0 & 0 & 0 & 0 & 0\\

e4 & \makecell{Purchase Requisition\\Line Created} & \makecell{2017-07-15\\15:00} & c1 & rq2 & & & & A457 & & & & & 0 & 0 & 0 &  & 0 & 0 & 0 & 0 & 0 & & 0 & 0 & 0 & 0 & 0 & 0 & 0\\

e5 & \makecell{Purchase Order\\Line Creation} & \makecell{2017-07-16\\15:00} & c1 & & o1 & & &
 A458 & 100 & 7 & 100\_L50 & & 1 & 0 & 0 & 100 & 1 & 0 & 1 & 0 & 0 & & 1 & 0 & 0 & 0 & 0 & 0 & 0\\

e6 & \makecell{Purchase Order\\Line Creation} & \makecell{2017-07-17\\15:00} & & rq1 & o2 & & &
 A458 & 200 & 8 & 100\_L51 & & 2 & 0 & 0 & 150 & 0.5 & 0.5 & 0.5 & 0.5 & 0 & & 1 & 0 & 0 & 0 & 0 & 0 & 0\\

e8 & \makecell{Goods Line\\Registered} & \makecell{2017-07-22\\15:00} & & & o1 & r1 & & A456 &
 100 & 7 & 100\_L50 & 10 & 2 & 1 & 0 & 150 & 0.5 & 0.5 & 0.5 & 0.5 & 0 & 10 & 1 & 0.5 & 1 & 0 & 0 & 0 & 0\\

e9 & \makecell{Invoice\\Receipt} & \makecell{2017-07-22\\16:00} & & & & & i1 & A125 & & & & & 2 & 1 & 1 & 150 & 0.5 & 0.5 & 0.5 & 0.5 & 0 & 10 & 1 & 0.5 & 1 & 1 & 0 & 0 & 0\\

e10 & \makecell{Purchase Requisition\\Group Changed} & \makecell{2017-07-22\\19:00} & & rq1 & & & & A456 & & & & & 2 & 1 & 1 & 150 & 0.5 & 0.5 & 0.5 & 0.5 & 0 & 10 & 1 & 0.5 & 1 & 1 & 0 & 0 & 0\\

e11 & \makecell{Purchase Order\\Line Creation} & \makecell{2017-07-23\\9:00} & & rq1 & o4 & & &
 A458 & 600 & 8 & 100\_L51 & & 3 & 1 & 1 & 300 & 0.33 & 0.66 & 0.33 & 0.66 & 0 & 10 & 1 & 0.33 & 1 & 1 & 0 & 0 & 0\\

e12 & \makecell{Goods Line\\Registered} & \makecell{2017-07-23\\15:00} & & & o2 & r2 & & A456 &
200 & 8 & 100\_L50 & 10 & 3 & 2 & 1 & 300 & 0.33 & 0.66 & 0.33 & 0.66 & 0 & 10 & 1 & 0.66 & 1 & 1 & 0 & 0 & 0\\

e13 & \makecell{Invoice\\Registered} & \makecell{2017-07-29\\11:00} & & & & r1,r2 & i1 & A125 &
 & & & 10 & 3 & 2 & 1 & 300 & 0.33 & 0.66 & 0.33 & 0.66 & 0 & 10 & 1 & 0.66 & 1 & 1 & 1 & 1 & 0\\

e14 & \makecell{Invoice\\Cleared} & \makecell{2017-07-30\\12:00} & & & & & i1 & A125 & & & & & 3 & 2 & 1 & 300 & 0.33 & 0.66 & 0.33 & 0.66 & 0 & 10 & 1 & 0.66 & 1 & 1 & 1 & 1 & 1\\

e15 & \makecell{Goods Line\\Registered} & \makecell{2017-07-31\\15:00} & & & o4 & r3 & & A456 &
600 & 8 & 100\_L51 & 10 & 3 & 3 & 1 & 300 & 0.33 & 0.66 & 0.33 & 0.66 & 0 & 10 & 1 & 1 & 1 & 1 & 1 & 1 & 1\\

e16 & \makecell{Invoice\\Registered} & \makecell{2017-08-10\\11:00} & & & & r2,r3 & i2 & A125 &
& & & 10 & 3 & 3 & 2 & 300 & 0.33 & 0.66 & 0.33 & 0.66 & 0 & 10 & 1 & 1 & 1 & 0.5 & 1 & 1 & 0.5\\

e17 & \makecell{Invoice\\Cleared} & \makecell{2017-08-15\\14:00} & & & & & i2 & A125 & & & & & 3 & 3 & 2 & 300 & 0.33 & 0.66 & 0.33 & 0.66 & 0 & 10 & 1 & 1 & 1 & 0.5 & 1 & 1 & 1\\

\hline

e4 & \makecell{Purchase Requisition\\Line Created} & \makecell{2017-07-15\\15:00} & c1 & rq2 & & & & A457 & & & & & 0 & 0 & 0 &  & 0 & 0 & 0 & 0 & 0 & & 0 & 0 & 0 & 0 & 0 & 0 & 0\\

e7 & \makecell{Purchase Order\\Line Creation} & \makecell{2017-07-18\\15:00} & & rq2 & o3 & & &
 A458 & 300 & 8 & 100\_L52 & & 1 & 0 & 0 & 300 & 0 & 1 & 0 & 0 & 1 & & 1 & 1 & 0 & 0 & 0 & 0 & 0\\

e18 & \makecell{Goods Line\\Registered} & \makecell{2017-08-16\\15:00} & & & o3 & r4 & & A456 & 300 & 8 & 100\_L52 & 5 & 1 & 1 & 0 & 300 & 0 & 1 & 0 & 0 & 1 & 5 & 1 & 1 & 1 & 0 & 0 & 0 & 0\\

e19 & \makecell{Purchase Requisition\\Supplier Changed} & \makecell{2017-08-16\\17:00} & & rq2 & & & & A456 & & & & & 1 & 1 & 0 & 300 & 0 & 1 & 0 & 0 & 1 & 5 & 1 & 1 & 1 & 0 & 0 & 0 & 0\\

e20 & \makecell{Invoice\\Registered} & \makecell{2017-08-18\\11:00} & & & & r4 & i3 & A125 &
& & & 5 & 1 & 1 & 1 & 300 & 0 & 1 & 0 & 0 & 1 & 5 & 1 & 1 & 1 & 0 & 1 & 1 & 0\\

e21 & \makecell{Invoice\\Cleared} & \makecell{2017-08-20\\14:00} & & & & & i3 & A125 & & & & & 1 & 1 & 1 & 300 & 0 & 1 & 0 & 0 & 1 & 5 & 1 & 1 & 1 & 0 & 1 & 1 & 0\\

\hline

\end{tabular}
}
\label{tab:event_log_our_approach_req_aggregated}
\end{sidewaystable}

\begin{example}
The result of introducing the \aggrAttrs to the feature set is presented in Table \ref{tab:event_log_our_approach_req_aggregated}:
\begin{itemize}
\item The first type of \aggrAttr that was introduced is represented by the column \texttt{Avg order\_price}, which represents the average order price (indicated by the column \texttt{order\_price}) considering all the orders associated to the selected requisition. It can be seen that event \texttt{e5} is associated to an \texttt{Avg order\_price} of 100, since there is only one order with \texttt{order\_price} 100; conversely, event \texttt{e6} is associated to an \texttt{Avg order\_price} of 150, since there is one order (o1) with \texttt{order\_price} 100 and one order (o2) with \texttt{order\_price} 200.
\item The second type of \aggrAttr that was introduced is represented by the column \texttt{\%order\_delivery\_month=7}, which indicates the fraction of objects correlated with objects of type order where attribute \texttt{order\_delivery}\linebreak\texttt{\_month} takes on value 7: in particular, it indicates the fraction of the orders that are delivered in July (month 7), compared to the total number of orders. As an example, the event \texttt{e5} is associated with a \texttt{\%order\_delivery\_month=7} of 1, since there is only one order with delivery in July; conversely, event \texttt{e6} is associated with a \texttt{\%order\_delivery\_month=7} of 0.5, since there is one order (o1) with delivery in July and one order (o2) with delivery in August (month 8).
\item The third type of \aggrAttr introduced is represented by the column \texttt{\#Orders}, which represents the number of objects of type order associated to the selected requisition. In the event \texttt{e5} there is only one order (o1) associated to the selected requisition, while in the event \texttt{e6} there are two orders (o1 and o2) associated to the selected requisition.
\item Finally, the fourth type of \aggrAttr introduced is represented by the column \texttt{Order, \%Goods Line Registered}, which indicates the percentage of orders that have performed the activity \texttt{Goods Line Registered} at least once; as an example, it has a value of 0.5 for event \texttt{e8}, since among the two orders (o1 and 02) there is only one order (o1) that has performed \texttt{Goods Line Registered}.
\end{itemize}
\label{example_aggr_attributes}
\end{example}

\section{Implementation and Experiments}
\label{sec:experiments}
Our approach described in Section~\ref{sec:framework description} have been implemented in Python.
Prediction models were built using Catboost (cf.\ Section~\ref{sec:LiteraturePrediction}).
Experiments were conducted on an event log that records the real executions of an object-centric process of an utility provider company in Italy. In particular, we defined five different KPIs of which to predict the values.
Section~\ref{sec:Dataset} introduces the case study
employed for our evaluation, while Section~\ref{sec:Results} compares and evaluates proposed predictive techniques in Object-Centric Processes.
In particular, we compare our approaches with the na\"ive adoption of techniques related to single-id process predictive analytics, where only events directly associated with one object type are considered.

\subsection{Domain Description}
\label{sec:Dataset}
The evaluation of our approach was performed on the object-oriented process described as working example in Section \ref{sec:Preliminary}.
The process is real and is being executed by a well-known Italian utility provider company, which is also one of the major energy companies in Europe. The company focuses on the production/extraction of electricity and gas and on their distribution in different parts of the world.
As mentioned in Section \ref{sec:Preliminary}, the overall process runs through the intertwining of processes related to five different object types.

\noindent Section~\ref{sec:framework description} mentioned that our techniques require defining a viewpoint: for the KPIs relevant for this case study, we mainly considered the \textit{Contract} object type as viewpoint, but we also considered the \textit{Order} and the \textit{Requisition} as alternative possible viewpoints. As discussed, if we consider for example the \textit{Contract} viewpoint, traditional predictive analytics based on single-id event logs would only use the events related to \textit{Contract} (the na\"ive approach). Vice versa, our techniques would also use the information of the objects related to Contract via bridge events, including the interaction itself.

Before applying the na\"ive and our approach, we performed a preprocessing.
In particular, we removed attributes whose values were missing in more than 80\% of the cases, or whose values were always the same among all cases.
In the pre-processing phase, we used domain knowledge and removed the attributes that were somehow duplicate. For instance, the log contains an attribute that refers to the order plant name, which is unique, and a second related to plant identifier: one of the two can be removed.
Finally, the large dimension of the  utility provider company is also reflected in the cardinality of some categorical attributes. For instance, the codes of the materials that are shipped all around the world (\textit{order\_material\_code}) are stored in an attribute that counts up to $4,179$ different values. To reduce the cardinality of the attributes with thousands of different values, we used the 80-20 rule, a.k.a.\ Pareto Principle~\cite{pareto}. Specifically, we kept the most frequent attribute's values that cover the 80\% of cases, labelling the remaining values as "other".

We obtained a different event log for each of the illustrated techniques (na\"ive approach, approach without aggregated features, approach with aggregated features), consisting of $12,537$, $99,065$, and $10,349$ cases for the \textit{Contract}, the \textit{Order} and the \textit{Requisition} viewpoints respectively.

In each experiment, two thirds of the traces were used for training, and the rest as test set. Splitting was done in a complete random fashion, assuming no concept drift in the dataset.
In training, a hyperparameter optimization was carried out, using 20\% of the training data for the optimization (validation set).

In our evaluation we considered several KPIs, which can be grouped in three categories. The first is the \textit{path time}, defined as the elapsed time between a defined source activity \textit{a} and the last occurrence of the selected target activity \textit{b},
while the second (\textit{pay delay}) refers to the average number of unforeseen additional days that are needed to receive the payment of the invoice compared to the expected invoice payment due date.
Finally, the third category refers to whether or not a certain activity or a certain condition (e.g. a late payment) is eventually going to occur in the future.
The first two KPI categories are defined over a numerical domain while the second one is boolean with \textsf{true} indicating the occurrence, and \textsf{false} the absence.

\subsection{Results for Predictive Analytics in Object-Centric Processes}
\label{sec:Results}

For the evaluation we focused on three different viewpoints (\textit{Contract}, \textit{Requisition} and \textit{Order}) and we considered the two approaches, namely with and without \aggrAttrs, which are described in Section \ref{sec:framework description}. 
Examples~\ref{example_viewpoint_events} and~\ref{example_aggr_attributes} illustrate the application of our approaches to this case study, when the chosen viewpoint is the \textit{requisition}.

The KPIs that have been considered are reported in Table \ref{tab:kpi_statistics}; in particular, Table \ref{tab:numerical_kpi_statistics} illustrates the numerical KPIs, while Table \ref{tab:categorical_kpi_statistics} illustrates the categorical KPIs.
Each numerical KPI refers to the elapsed time from the first occurrence of the considered object to the last occurrence of a different selected activity.
The first five KPIs build on the contract viewpoint; in particular, the first KPI refers to the elapsed time from the creation of the contract to the last \textit{SES Line Registered}; this activity indicates that the service requested by the customer has been provided but, since the customer can require several services, it is of interest to know when all the services request have been provided.
The second KPI (\textit{SES Line Released}) indicates that a further step has been performed, which is the confirmation from the manager that everything was received correctly.
Another interesting KPI to be monitored is the elapsed time from the creation of the contract to the last \textit{Invoice Receipt}, activity that indicates that the invoice has been correctly charged to the customer; conversely, \textit{Invoice Cleared} indicates that the invoice has been paid.
The fifth KPI refers to the number of days exceeding the planned payment date (\textit{Pay Delay estimation}), considered starting from the creation of the contract to the last occurrence of \textit{Invoice Cleared}.
Finally, the remaining numerical KPIs refer to the elapsed time from the creation of the order or requisition to the last occurrence of some selected activity.

Moreover, after selecting the path from the creation of the contract to the last occurrence of \textit{Invoice Cleared}, we also considered three categorical KPIs, which were related to activities that the company wants to be prevented; in particular, it was interesting to know in advance whether there would be changes to the payment method (represented by the activity \textit{Invoice Pay Method Changed}): when this happens, usually there are delays with payments.
Furthermore, the company was also interested in forecasting whether there will be problems with the order (represented by the activity \textit{Purchase Order Blocked}), since this situation can bring additional delays caused by the reworks needed to fix the problem.
Finally, it was interesting to know whether there will be delays with the payments (represented by the attribute \textit{Pay Type} assuming value Late).

\begin{table} [t!]
\centering
\caption{Statistics related to the selected KPIs}
\begin{subtable}[c]{\textwidth}
    \caption{Numerical KPIs statistics}
    \footnotesize
    \centering
    \resizebox{\textwidth}{!}
    {\begin{tabular}{|l|l|l|l|l|l|l|l|}
    \hline
    \cellcolor{LightCyan}\textbf{KPI} &
    \cellcolor{LightCyan}\textbf{Viewpoint} &
    \cellcolor{LightCyan}{\textbf{\begin{tabular}[c]{@{}c@{}}Average value \\(days)\end{tabular}}} &
    \cellcolor{LightCyan}{\textbf{\begin{tabular}[c]{@{}c@{}}Standard deviation \\value (days)\end{tabular}}}\\
    \hline

    \textbf{Elapsed Time from Contract to the last SES Line Registered} &
    Contract &
    291.59 &
    224.17 \\  \hline

    \cellcolor{LightGray}\textbf{Elapsed Time from Contract to the last SES Line Released}  &
     \cellcolor{LightGray}Contract &
     \cellcolor{LightGray}292.73 &
     \cellcolor{LightGray}224.2\\ \hline

     \textbf{Elapsed Time from Contract to the last Invoice Receipt} &
     Contract &
     257.04 &
     227.41 \\
    \hline

    \cellcolor{LightGray}\textbf{Elapsed Time from Contract to the last Invoice Cleared}  &
    \cellcolor{LightGray}Contract &
    \cellcolor{LightGray}312.3 &
    \cellcolor{LightGray}240.02\\ \hline

    \textbf{Pay Delay estimation from Contract to the last Invoice Cleared} & Contract & 9.71 & 43.2\\ \hline

    \cellcolor{LightGray}\textbf{Elapsed Time from Order to the last Invoice Receipt} &
    \cellcolor{LightGray}Order &
    \cellcolor{LightGray}24.66 &
    \cellcolor{LightGray}28.08 \\  \hline

    \textbf{Elapsed Time from Order to the last Invoice Cleared}  & Order & 64.67 & 43.45\\ \hline

    \cellcolor{LightGray}\textbf{Elapsed Time from Requisition to the last Invoice Receipt} &
    \cellcolor{LightGray}Requisition &
    \cellcolor{LightGray}48.17 &
    \cellcolor{LightGray}42.79 \\  \hline

    \textbf{Elapsed Time from Requisition to the last Invoice Cleared}  & Requisition & 114 & 48.56\\ \hline

    \cellcolor{LightGray}\textbf{Elapsed Time from Requisition to the last SES Line Released} &
    \cellcolor{LightGray}Requisition &
    \cellcolor{LightGray}35.71 &
    \cellcolor{LightGray}40.72 \\  \hline

    \textbf{Elapsed Time from Requisition to the last SES Line Registered} & Requisition & 34.59 & 40.71\\ \hline

    \end{tabular}}
    \label{tab:numerical_kpi_statistics}
    \vspace{5mm}
\end{subtable}

\begin{subtable}[c]{\textwidth}
    \caption{Categorical KPIs statistics. Column \textit{\% cases} represents the percentage of cases in which the activity or the attribute was present with that particular value. Ideally it should be the lowest possible}
    \footnotesize
    \centering
    \resizebox{\textwidth}{!}
    {\begin{tabular}{|l|l|l|l|l|l|l|}
    \hline
    \cellcolor{LightCyan}\textbf{KPI} &
    \cellcolor{LightCyan}\textbf{Viewpoint} &
    \cellcolor{LightCyan}\textbf{\% cases}\\
    \hline

    \textbf{Occurrence of Activity Pay Method Changed (from Contract to the last Invoice Cleared)} &
    Contract &
    29\% \\  \hline

    \cellcolor{LightGray}\textbf{Occurrence of Activity Purchase Order Blocked (from Contract to the last Invoice Cleared)}  &
    \cellcolor{LightGray}Contract &
    \cellcolor{LightGray}27\%\\ \hline

    \textbf{Occurrence of Attribute Pay Type assuming value Late (from Contract to the last Invoice Cleared)} &
    Contract &
    52\% \\  \hline

    \end{tabular}}
    \label{tab:categorical_kpi_statistics}
    \vspace{5mm}
\end{subtable}
\label{tab:kpi_statistics}
\end{table}

Since eleven KPIs were numerical and the values were reasonably well balanced, we adopted the Mean Absolute Error (MAE).
The last three KPIs considered were instead categorical and related to activities that the company wants to be prevented; therefore, we computed the F1 score.
Since the selection of the cases and the overall duration of the case itself vary depending on the selected KPI, also the observed average duration and standard deviation of the cases are different; the statistics for the selected numerical KPIs, shown in Table \ref{tab:numerical_kpi_statistics}, are necessary to understand if our predictive model is achieving a good prediction quality.
Conversely, for the last three categorical KPIs, we reported in Table \ref{tab:categorical_kpi_statistics} on the distribution of the classes; in particular, the activity \textit{Invoice Pay Method Changed} was performed in the 29\% of the cases, the activity \textit{Purchase Order Blocked} was performed in the 27\% of cases, while the attribute \textit{Pay Type} appeared with value \textit{Late} in the 52\% of cases.

\begin{table*}[t!]
\caption{Prediction performances for the considered KPIs using the proposed techniques, measured in terms of Mean Absolute Error (MAE) or F1 score. The average predictions performances were obtained considering two different divisions of the cases in train and test sets. Training times are reported in brackets.}
\footnotesize
\centering
\resizebox{\textwidth}{!}
{\begin{tabular}{|l|l|l|l|l|l|}
\hline
\cellcolor{LightCyan}\textbf{Viewpoint} &
\cellcolor{LightCyan}{\textbf{\begin{tabular}[c]{@{}c@{}}KPI\end{tabular}}} &
\cellcolor{LightCyan}{\textbf{\begin{tabular}[c]{@{}c@{}}Na\"ive approach\end{tabular}}} &
\cellcolor{LightCyan}{\textbf{\begin{tabular}[c]{@{}c@{}}approach w.out\\ aggr. features\end{tabular}}} &
\cellcolor{LightCyan}{\textbf{\begin{tabular}[c]{@{}c@{}}approach w.\\ aggr. features\end{tabular}}} \\
\hline
 \textbf{Contract} &
 Elapsed Time from Contract to the last Invoice Receipt (MAE) &
 40.67 (5m) &
 31.72 (36m) &
 \textbf{28.49} (53m)\\
\hline
 \cellcolor{LightGray}\textbf{Contract}  &
 \cellcolor{LightGray}Elapsed Time from Contract to the last SES Line Released (MAE) &
 \cellcolor{LightGray}38.63 (7m) &
 \cellcolor{LightGray}29.96 (25m) &
 \cellcolor{LightGray}\textbf{26.71} (15m)\\ \hline

\textbf{Contract}  & Elapsed Time from Contract to the last Invoice Cleared (MAE)  & 44.73 (5m) & 37 (33m) & \textbf{33.06} (81m) \\ \hline

\cellcolor{LightGray}\textbf{Contract}  &
\cellcolor{LightGray}Elapsed Time from Contract to the last SES Line Registered (MAE)  &
\cellcolor{LightGray}39.65 (5m)&
\cellcolor{LightGray}30.03 (24m)&
\cellcolor{LightGray}\textbf{28.81} (28m)\\  \hline

\textbf{Contract}  & Activity Invoice Pay Method Changed occurrence (F1)  & 0.83 (12m) & 0.86 (78m) & \textbf{0.87} (95m)\\ \hline

\cellcolor{LightGray}\textbf{Contract}  &
\cellcolor{LightGray}Activity Purchase Order Blocked occurrence (F1) &
\cellcolor{LightGray}0.69 (12m) &
\cellcolor{LightGray}0.71 (76m) &
\cellcolor{LightGray}\textbf{0.74} (94m)\\  \hline

\textbf{Contract}  & Attribute Pay Type Late occurrence (F1)  & 0.88 (12m) & 0.88 (78m) & \textbf{0.89} (95m)\\ \hline

\cellcolor{LightGray}\textbf{Contract}  &
\cellcolor{LightGray}Pay Delay estimation from Contract to the last Invoice Cleared (MAE)  &
\cellcolor{LightGray}15.01 (5m) &
\cellcolor{LightGray}14.09 (37m) &
\cellcolor{LightGray}\textbf{13.6} (29m)\\  \hline

\textbf{Order}  & Elapsed Time from Order to the last Invoice Receipt (MAE)  & 11.62 (24m) & \textbf{9.50} (3h 30m) & 10.39 (3h)\\ \hline

\cellcolor{LightGray}\textbf{Order}  &
\cellcolor{LightGray}Elapsed Time from Order to the last Invoice Cleared (MAE)  &
\cellcolor{LightGray}15.62 (19m)&
\cellcolor{LightGray}12.03 (11h)&
\cellcolor{LightGray}\textbf{11.51} (10h)\\  \hline

\textbf{Requisition}  & Elapsed Time from Requisition to the last Invoice Receipt (MAE)  & 17.65 (7m) & \textbf{11.43} (31m) & 11.7 (33m) \\ \hline

\cellcolor{LightGray}\textbf{Requisition}  &
\cellcolor{LightGray}Elapsed Time from Requisition to the last Invoice Cleared (MAE)  &
\cellcolor{LightGray}21.09 (6m)&
\cellcolor{LightGray}14.53 (38m)&
\cellcolor{LightGray}\textbf{12.33} (44m)\\  \hline

\textbf{Requisition}  & Elapsed Time from Requisition to the last SES Line Released (MAE) & 17.56 (5m) & 12.84 (11m) & \textbf{12.79} (12m) \\ \hline

\cellcolor{LightGray}\textbf{Requisition}  &
\cellcolor{LightGray}Elapsed Time from Requisition to the last SES Line Registered (MAE)  &
\cellcolor{LightGray}17.35 (5m)&
\cellcolor{LightGray}\textbf{11.3} (14m)&
\cellcolor{LightGray}13.08 (14m)\\  \hline

\end{tabular}}
\label{tab:prediction_performances}
\end{table*}

Results of proposed predictive techniques on the selected KPIs are shown in Table~\ref{tab:prediction_performances}; in particular, they represent the average accuracy obtained considering two different random splits of the cases in train and test sets.
Compared to the na\"ive approach, it can be seen that the two approaches that consider the object-interaction show improved predictive performances in all the considered KPIs.
The additional use of aggregated attributes feeds in additional information that enables a further improvement of the prediction quality in a vast majority of cases.

Table \ref{tab:prediction_performances} also reports on the training time needed to train the prediction model of the na\"ive and our approaches on each KPI of interest.
As it can be seen, our approaches have the highest training time in all KPIs. Note that this does not pose significant limitations, since it is just performed once; after the model has been trained, all the predictions on the test set can be obtained in less than one second for all the considered KPIs.

\begin{figure}[t!]
    \centering
    \includegraphics[width=\textwidth]{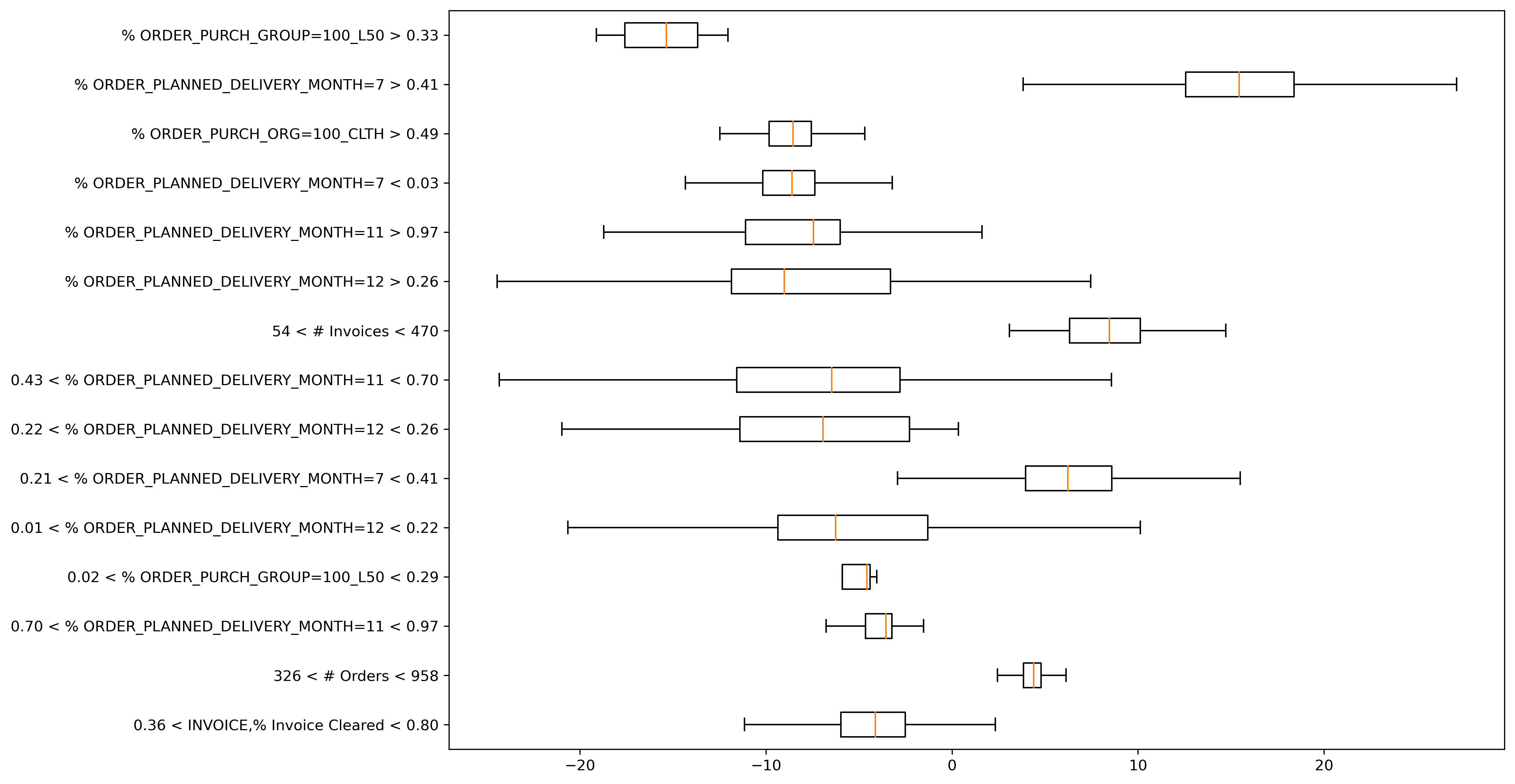}
    \caption{Boxplot representing the impact of some of the most important aggregated features that were added to the predictive model in order to improve the accuracy for the KPI \textit{Elapsed Time from Contract to the last Invoice Receipt} in the approach with aggregated features.}
    \label{inv_rec_aggr_boxplot}
\end{figure}

As a further confirmation of the importance of object-interaction and aggregated features, we computed the Shapley Values in accordance to the framework discussed in Section \ref{sec:explanations}; in particular, explanations were calculated on the test dataset.
Figure~\ref{inv_rec_aggr_boxplot} illustrates the boxplot representing the distribution of the Shapley values over the different trace prefixes for some of the most important features influencing the prediction of the KPI \textit{Elapsed Time from Contract to the last Invoice Receipt} for the approach with the aggregated features. In particular, each boxplot is ordered by the average Shapley value on the selected KPI considering the absolute value. Each row of the boxplot is linked to an explanation, which extends towards left or right, depending whether the observed Shapley values for the explanation were negative or positive.
It can be clearly seen in Figure~\ref{inv_rec_aggr_boxplot} that many aggregated features are indeed considered relevant by the predictive model; as an example, the most important explanation is \textit{\% ORDER\_PURCH\_GROUP=100\_L50 $>$ 0.33} and the average associated Shapley value is -15 days: this means that, when more than the 33\% of the orders are related to the purchase group \textit{100\_L50}, the estimated Elapsed Time from the signature of the Contract to the last Invoice Receipt reduces on average by 15 days.

\begin{figure}[t!]
    \centering
    \includegraphics[width=\textwidth]{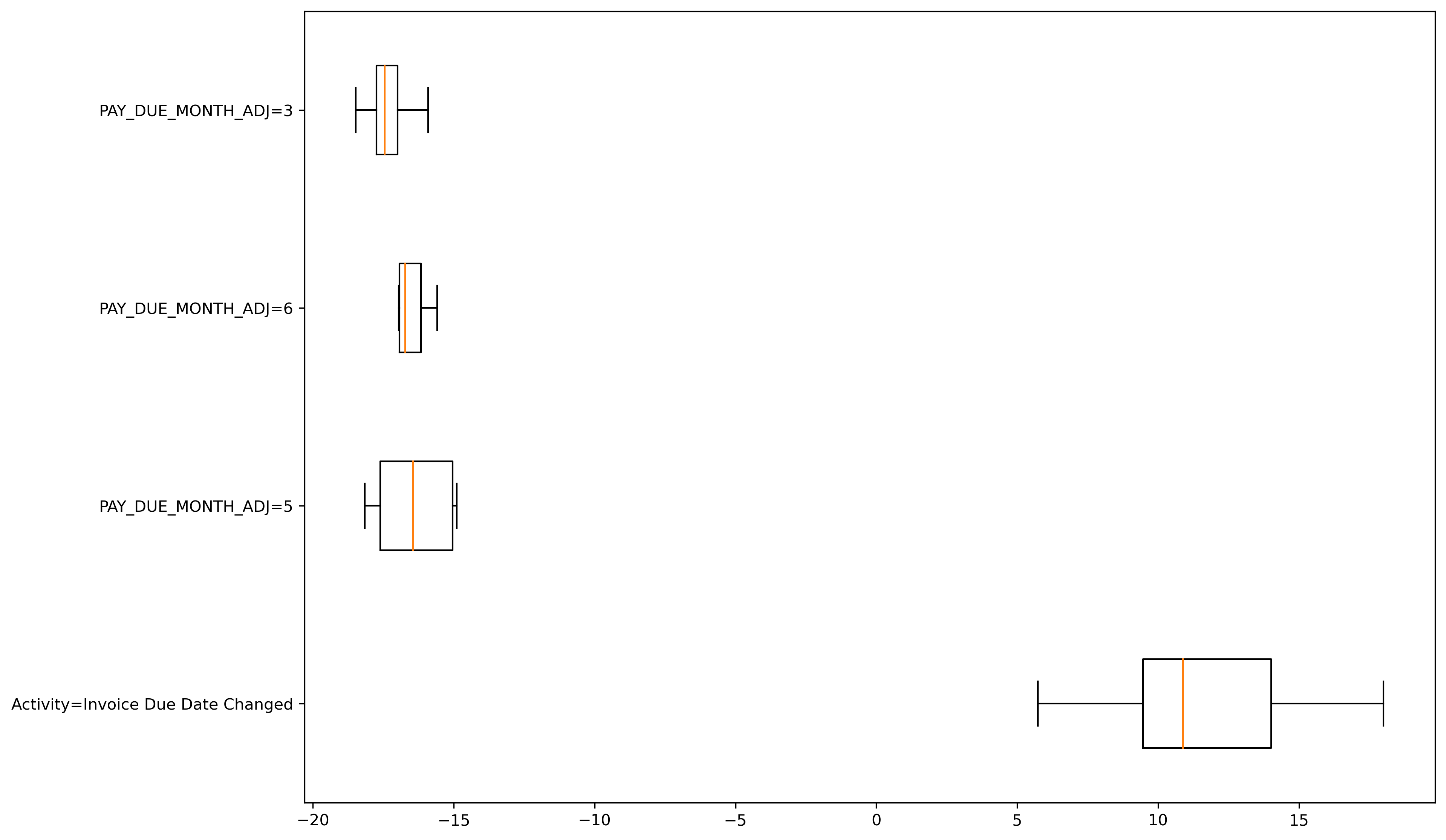}
    \caption{Boxplot representing the impact of some of the most important features that were added to the predictive model (such as the attributes related to the \textit{Invoice} object type) in order to improve the accuracy for the KPI \textit{Elapsed Time from Contract to the last Invoice Cleared} in the approach without aggregated features.}
    \label{inv_cl_no_aggr_boxplot}
\end{figure}

\begin{figure}[t!]
    \centering
    \includegraphics[width=\textwidth]{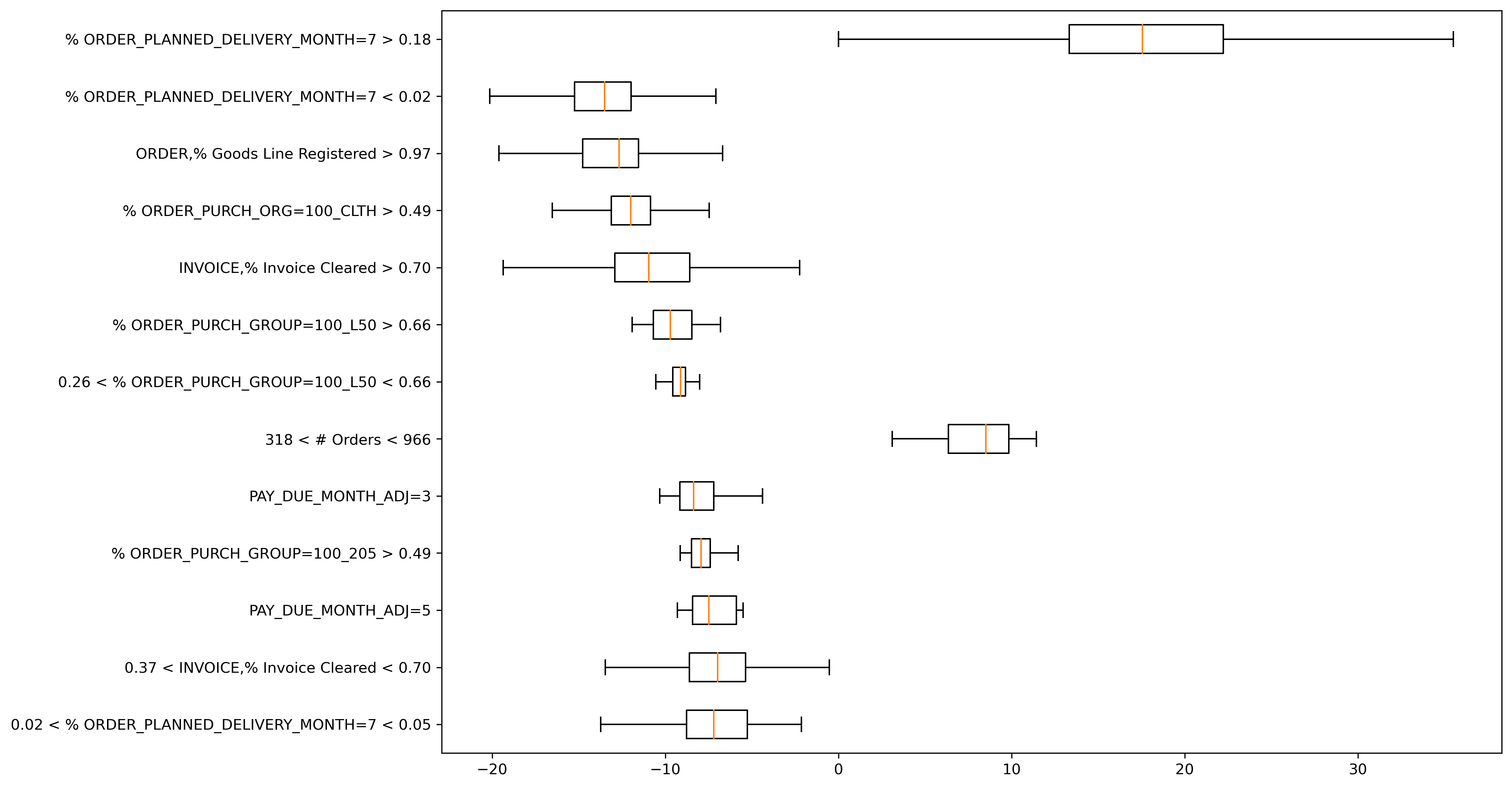}
    \caption{Boxplot representing the impact of some of the most important aggregated features that were added to the predictive model in order to improve the accuracy for the KPI \textit{Elapsed Time from Contract to the last Invoice Cleared} in the approach with aggregated features.}
    \label{inv_cl_aggr_boxplot}
\end{figure}

A similar reasoning can be applied to Figure~\ref{inv_cl_no_aggr_boxplot},
which reports on some of the most important features influencing the prediction of the KPI \textit{Elapsed Time from Contract to the last Invoice Cleared} for the approach without the aggregated features.
As it can be seen, several features were added after considering the object interaction and were considered useful by the predictive model, such as \textit{PAY\_DUE\_MONTH\_ADJ} and the occurrence of the activity \textit{Invoice Due Date Changed}; these are attributes related to the \textit{Invoice} object type, whose values are related to the last opened invoice (or to the last performed activity).

However, since we do not use aggregated features, we miss several factors significantly influencing the prediction. Indeed, when we consider the aggregated features for the same KPI (\textit{Elapsed Time from Contract to the last Invoice Cleared}) and compute the significance of the influence (i.e. the average Shapley value), we obtain the boxplots in Figure~\ref{inv_cl_aggr_boxplot}.
It can be clearly seen that the \aggrAttrs are now among the most important factors significantly influencing the prediction. 
As an example, the two most important factors that are contributing to increase the estimated time are represented by \textit{\%Order\_planned\_delivery\_month=7 $>$ 0.18} and \textit{318 $<$ \#Orders $<$ 966}. The former is associated with an average Shapley value of 20, which means that when more than 18\% of the orders are delivered in July (month 7), the estimated time to clear the last invoice is 20 days larger that the average. The latter, associated with an average Shapley value of 8, indicates that when there are a lot of orders that needs to be managed at the same time (between 318 and 966), the estimated time to clear the last invoice is 8 days larger that the average.
Conversely, one of the most important factors contributing to decrease the estimated time is  
\textit{ORDER, \%Goods Line Registered $>$ 0.97}, which is associated with an average Shapley value of -10; this means that, when the activity Goods Line Registered (which represents the fact that the goods have been received) has been performed at least once in more than 97\% of the orders, the estimated time reduces on average by 10 days.

\section{Related Works}
\label{sec:related}

A body of research exists on object-centric processes. Several research works focus on modelling object-centric processes (e.g.~\cite{Hull}) and the verification of the correctness of these models \cite{10.1145/2661829.2662050,M_C@IJSTTT16}.
In the realm of Process Mining, techniques are proposed to discover object-centric process models and behavioral dependencies between objects (i.e.\ artifacts) \cite{vdA_B@FundInformaticae20,7229358,10.1007/978-3-319-06257-0_3,8010712}, and to tackle the problem of the object-centric process conformance checking \cite{10.1007/978-3-642-23059-2_26,9576886}.
However, none of the existing works consider object-centric process predictive analytics.
A few works consider interactions among different instances of a process \cite{Senderovich2019FromKT,DBLP:conf/icpm/DenisovFA19,DBLP:conf/icpm/KlijnF20}, but they still rely on the notion of a single process flow (i.e., single case identifier). While some of these works provide valuable insights into inter-case features, their extension to object-centric process is in fact the goal of the technique proposed in this framework.

The work proposed in this paper is based on the idea to unfold an object-centric event log into traditional events logs around a viewpoint. Berti et al.~\cite{B_vdA@SIMPDA19} have introduced a very similar concept of viewpoint to indeed extract event logs from the data stored in relational databases: however, the concept of viewpoint is not aimed at process predictive analytics. In fact, unfolding typically leads to a duplication of events, possibly strengthening existing directly-follows relationships or, even, adding new relationships that do not exist in reality. These problems are known as convergence and divergence \cite{10.1007/978-3-030-30446-1_1,DBLP:journals/jodsn/EsserF21} and make it impossible to apply process-mining techniques designed for single-flow processes that heavily rely on the directly-follow relationships, such as those for model discovery and conformance checking. Conversely, unfolding causes no problem in our process-prediction approach, because we do not use the direct-follow relationships.

\section{Conclusions}
\label{sec:conclusions}

The lion's share of attention in Business Process Management (BPM) has traditionally been on designing and analyzing processes that are based on a unique notion of case identifier, with a single flow of execution from an initial state to one of the potential final states.
In practice, organizations execute more complex processes that are often interacting with each other: one instance of a given process synchronizes with instances of other processes, possibly exchanging data.
In light of above, the object-centric process paradigm is nowadays attracting more and more attention in academic and industry. In this paradigm, the process is seen as the interplay of numerous sub-processes that constitute life cycles of different objects of various types, where these life cycles period synchronize with each other.

This paper tackles the problem of predictive analytics over object-centric processes. The large share of research in predictive analytics cannot be directly applied here, because it traditionally focuses on the problem of predicting the outcome of cases (i.e., process instances) that run in isolation. Also recent techniques that capture the inter-case dynamics assume instances to be of the same process, namely referring to the same object type. If the valuable information of the interaction between objects of the same or different type is not fed into the construction of prediction models, the resulting model might be of low accuracy.

This paper reports on an approach where object-centric logs are unfolded into traditional event logs around a viewpoint, namely around a process' object type. This enables leveraging on the-state-of-art techniques for process predictions.

We conducted experiments with an event log related to a real object-centric process being executed by an utility company in Italy, and measured the accuracy of the predictions using our approach. The accuracy was subsequently also compared with that of a na\"ive approach that only considers the events related to the process of the viewpoint objects. Experiments have shown that the na\"ive approach performs rather poorly, confirming the importance of our approach to consider the object interactions when predicting.

As future work, we plan to test on other datasets. We also aim to leverage on prediction techniques based on Graph Neural Networks~\cite{4700287,DBLP:conf/icml/WuSZFYW19}, where the relationships between object types can be explicitly represented as, e.g., a UML or ER diagram (cf.\ Figure~\ref{fig:Uml_diagram}). This would likely provide further information for higher-quality predictions.

\bibliography{references}

\end{document}